\documentclass{article}

\usepackage{microtype}
\usepackage{graphicx}
\usepackage{subfigure}
\usepackage{booktabs} 

\usepackage{hyperref}
\usepackage{multirow}

\usepackage[accepted]{icml2025}

\usepackage{amsmath}
\usepackage{amssymb}
\usepackage{mathtools}
\usepackage{amsthm}

\usepackage[capitalize,noabbrev]{cleveref}

\theoremstyle{plain}

\theoremstyle{definition}

\theoremstyle{remark}

\usepackage[textsize=tiny]{todonotes}

\icmltitlerunning{(GG) MoE vs. MLP on Tabular Data}

\begin{document}

\twocolumn[
\icmltitle{(GG) MoE vs. MLP on Tabular Data}



\icmlsetsymbol{equal}{*}

\begin{icmlauthorlist}
\icmlauthor{Andrei Chernov}{yyy}
\end{icmlauthorlist}

\icmlaffiliation{yyy}{Independent Researcher}

\icmlcorrespondingauthor{Andrei Chernov}{chernov.andrey.998@gmail.com}

\icmlkeywords{Machine Learning, ICML, Muxture of Experts, Deep Learning, Tabular Data}

\vskip 0.3in
]



\printAffiliationsAndNotice{}  

\begin{abstract}
In recent years, significant efforts have been directed toward adapting modern neural network architectures for tabular data. However, despite their larger number of parameters and longer training and inference times, these models often fail to consistently outperform vanilla multilayer perceptron (MLP) neural networks. Moreover, MLP-based ensembles have recently demonstrated superior performance and efficiency compared to advanced deep learning methods. Therefore, rather than focusing on building deeper and more complex deep learning models, we propose investigating whether MLP neural networks can be replaced with more efficient architectures without sacrificing performance. In this paper, we first introduce GG MoE, a mixture-of-experts (MoE) model with a Gumbel-Softmax gating function. We then demonstrate that GG MoE with an embedding layer achieves the highest performance across $38$ datasets compared to standard MoE and MLP models. Finally, we show that both MoE and GG MoE utilize significantly fewer parameters than MLPs, making them a promising alternative for scaling and ensemble methods.
\end{abstract}

\section{Introduction}
\label{intro}
Supervised machine learning on tabular data is widely applied, and its business value is undeniable, leading to the development of numerous algorithms to address these problems. Gradient Boosting Decision Tree (GBDT) models \cite{chen2016xgboost,ke2017lightgbm,prokhorenkova2018catboost} have demonstrated superior performance compared to deep learning methods \cite{shwartz2022tabular,grinsztajn2022tree} and remain the most common and natural choice for tabular data prediction. As a result, tabular data remains one of the few domains where neural networks do not yet dominate.

In recent years, many researchers have attempted to adapt transformer-based neural network architectures for tabular data \cite{huang2020tabtransformer,somepalli2021saint,song2019autoint}. While these approaches have shown promising results on specific subsets of datasets, they often fail to consistently outperform vanilla Multilayer Perceptron (MLP) neural networks across a wide range of datasets \cite{gorishniy2024tabm}. This is despite their significantly higher computational requirements and larger number of parameters. Furthermore, a study \cite{gorishniy2024tabm} demonstrated that efficient ensembles of MLPs tend to outperform advanced deep learning models. 

This raises a question that, in our view, has been overlooked in recent research: Is there a neural network architecture that is more efficient than MLPs in terms of parameter count while still achieving comparable performance? 

Framing the problem this way makes investigating the performance of Mixture-of-Experts (MoE) models on tabular data a natural choice. MoE has recently gained popularity in deep learning \cite{fedus2022review}. However, to the best of our knowledge, little research has explored the adaptation of MoE to the tabular deep learning domain or evaluated its performance across a broad range of datasets. In this paper, we aim to address this gap. 

Additionally, we introduce MoE with a Gumbel-Softmax activation function on the output of the gating network (GG MoE); see \cref{gg_moe} for details. We compare the performance of MoE and GG MoE with MLP across 38 datasets and demonstrate that GG MoE achieves the highest average performance while both MoE and GG MoE are significantly more parameter-efficient than MLP.

\section{Related Work}
\label{related}

\subsection{Tabular Deep Learning}
Although it is theoretically proven that feedforward neural networks can approximate functions from a wide family with arbitrary accuracy \cite{cybenko1989approximation,hornik1991approximation}, in practice, they often underperform compared to GBDT methods in the tabular domain \cite{gorishniy2021revisiting,shwartz2022tabular,grinsztajn2022tree}. To improve performance, extensive research has been conducted. Here, we highlight three main directions.

The first direction focuses on improving feature preprocessing \cite{gorishniy2022embeddings,guo2021embedding} or enhancing the training process \cite{bahri2021scarf,gorishniy2021revisiting,jeffares2023tangos,holzmuller2024better,kadra2021well}. The second direction attempts to adapt more advanced neural network architectures, such as transformer-based models \cite{huang2020tabtransformer,somepalli2021saint,song2019autoint}. Although these architectures show promising results on specific datasets, they often fail to consistently outperform vanilla MLPs across a wide range of datasets while requiring significantly more computational resources \cite{gorishniy2024tabm}. 

The third line of research explores neural network ensembles, which generate multiple predictions for each data point and aggregate them into a final scalar prediction. The most straightforward ensembling approach involves training multiple neural networks independently and averaging their results \cite{lakshminarayanan2017simple}. While this improves performance, it demands significantly more computational resources. Recent studies have investigated more efficient ensembling methods, such as partially sharing weights across different neural networks \cite{gorishniy2024tabm,wen2020batchensemble}. Although ensembles with shared weights tend to improve performance, they still require significantly more computational resources than GBDT and MLP models \cite{gorishniy2024tabm}. In this paper, we investigate more efficient architectures.

\subsection{Mixture of Experts}
Mixture of Experts (MoE) is not a new architecture, and extensive research has been conducted on it. We encourage readers to refer to the comprehensive survey by \cite{yuksel2012twenty} for an overview. MoE consists of two main components: a gating function and expert functions. The experts can be considered an ensemble of different models, which are aggregated into a final prediction using a gating function.\footnote{While a gating function and expert functions do not necessarily have to be neural networks, in this paper, we assume that they are neural networks when referring to MoE.} A detailed description of the MoE architecture is provided in \cref{moe}.

MoE was not a widely adopted choice in deep learning architectures until its recent application to natural language processing (NLP) \cite{du2022glam} and computer vision (CV) \cite{puigcerver2023sparse,riquelme2021scaling}. Various MoE architectures have been developed and tested for these domains \cite{fedus2022review}. However, to the best of our knowledge, few studies have evaluated the performance of MoE across a broad range of tabular datasets. In this paper, we aim to address this gap.

\subsection{Gumbel-Softmax Distribution}
The Gumbel-Softmax distribution is widely used in deep learning for its ability to produce differentiable samples that approximate non-differentiable categorical distributions \cite{jang2016categorical}. In this paper, we utilize this distribution for a different purpose—primarily to regularize the gating neural network in MoE (see \cref{gg_moe}).

\section{Models}
In this paper, we compare the performance of three models: MLP, MoE, and GG MoE. In this section, we provide a brief introduction to the architecture of each.

\subsection{Notation}
We formulate the supervised machine learning problem using a probabilistic framework. Given a training dataset consisting of $N$ independent and identically distributed (i.i.d.) observations $x_i \in \mathbb{R}^M$, where $i = 1, \ldots, N$ and $M$ is the input dimension, along with their corresponding target values $y_i$, our goal is to model the conditional distribution $p(y \mid x, w)$.

\subsection{MoE}
\label{moe}
Informally, MoE consists of two main components. The first component comprises $K$ experts, which are independent models that learn the target distribution. These experts do not share weights and can be executed in parallel. The second component is a gating function, which maps an input to a probability distribution over the experts. The final output is a weighted average of the experts' outputs, where the weights are determined by the gating function. The Deep Ensemble method from \cite{lakshminarayanan2017simple} can be considered a special case of an MoE model with a constant gating function, $g = 1/K$.  

More formally, the target distribution is defined as:

\begin{equation}
p(y \mid x, w) = \sum_{i=1}^K p(i \mid x, w_g) p(y \mid i, x, w_{e_i}),
\end{equation}

where $K$ is the number of experts, and $p(i \mid x, w_g)$ is modeled by the gating function $g$. To model $g$, we use a multiclass logistic regression\footnote{To simplify notation and include a bias term, we use the notation $[x,1]$, which means we artificially add a constant feature equal to $1$ as the last dimension of each data point, leading to $w_g \in \mathbb{R}^{K \times (M+1)}$.}:

\begin{equation}
\begin{split}
g(i \mid x, w_g) &= \text{softmax}_i(w_g^T [x,1]) = \\
    &= \frac{\exp(w_{g_i}^T [x,1])}{\sum_{j=1}^K \exp(w_{g_j}^T [x,1])}.
\end{split}
\end{equation}

Finally, $p(y \mid i, x, w_{e_i})$ is modeled by the $i$-th expert, which, in our case, is an MLP neural network, as described in \cref{mlp}.

\subsection{MLP}
\label{mlp}

For an MLP neural network, we model the target distribution as  
\begin{equation}
p(y \mid x, w) = \delta(y - f(x;w))
\end{equation}
for regression tasks, where $\delta$ is the Dirac delta function. For classification tasks, we define  
\begin{equation}
p(y = i \mid x, w) = \text{softmax}_i(f(x;w)).
\end{equation}
Here, $f(x, w)$ represents the model function, which, in our case, is a neural network parameterized by $w$, mapping an input observation $x$ to a target value $y$.  

The model function $f(x; w)$ consists of a sequence of $n$ blocks followed by a final linear layer. Each block is defined as  
\begin{equation}
\text{Block}_i = \text{Dropout}(\text{ReLU}(\text{Linear}(x;w_i))),
\end{equation}
where $w_i$ represents the parameters of the $i$-th block. All linear layers within the blocks share the same hidden dimension. The final linear layer produces an output of dimension $1$ for regression problems or $C$ for classification tasks, where $C$ is the number of classes.  

An MLP neural network can be considered a degenerate case of MoE with a single expert and a constant gating function, $g=1$.

\subsection{GG MoE}
\label{gg_moe}
The only difference between MoE and GG MoE is that, instead of the standard softmax function, we use the Gumbel-Softmax function in the gating mechanism:
\begin{equation}
g_G(i \mid x, w_g) = \frac{\exp\left(\frac{w_{g_i}^T [x,1] + s_i}{\tau}\right)}{\sum_{j=1}^K \exp\left(\frac{w_{g_j}^T [x,1]+ s_j}{\tau}\right)},
\label{eq:softmax_adjusted}
\end{equation}
where $s_1, s_2, \ldots, s_K$ are i.i.d. samples drawn from the $\text{Gumbel}(0,1)$ distribution. As $\tau \to +\infty$, the Gumbel-Softmax distribution converges to a uniform distribution, while as $\tau \to 0$, it converges to an argmax distribution. Due to this property, Gumbel-Softmax has been widely used in deep learning to sample from discrete distributions \cite{jang2016categorical}. However, in this paper, we utilize this distribution primarily for regularization purposes.  

It is well known that, without intervention during training, the gating function may converge to a degenerate distribution, where one expert receives a weight close to $1$, while all others receive weights close to $0$. The authors of \cite{shazeer2017outrageously} proposed adding Gaussian noise to the softmax operation to prevent this behavior. In our research, we prefer Gumbel noise over Gaussian noise or other alternatives because, in our view, Gumbel-Softmax exhibits more suitable asymptotic behavior for this role. Specifically, as $\tau \to 0$, the gating output distribution converges to an argmax distribution, leading to the entropy  
$h(p) = -\sum_{i = 1}^K p_i \log p_i$
approaching zero. Conversely, as $\tau \to +\infty$, the distribution becomes uniform, attaining the highest possible entropy value of $\log(K)$. For details on entropy and its properties, we refer to \cite{conrad2004probability}. Informally, entropy can be interpreted as a measure of uncertainty. We tune $\tau$ as a hyperparameter for each dataset (see \cref{hyperparameters}), effectively selecting the "optimal level of uncertainty."  

A drawback of introducing stochasticity into the gating function is the challenge of handling it during inference. To obtain an unbiased estimation of the target distribution, we apply Monte Carlo (MC) estimation \cite{graham2013stochastic} to approximate the expected value:
\begin{equation}
\label{monte_carlo}
\begin{split}
    E[y(x; w)] &= \sum_{i=1}^K g_G(i \mid x, w_g) f(x; w) \approx \\
    &\approx \frac{1}{N} \sum_{j=1}^N \sum_{i=1}^K \alpha_{ji} f(x; w),
\end{split}
\end{equation}
where $\alpha_j$ are i.i.d. samples from the Gumbel-Softmax distribution. This estimation introduces a minor runtime overhead during inference. Fortunately, the sampling procedure is computationally inexpensive, as we do not need to recalculate logits or expert predictions for different samples. As shown in \cref{sec:results}, $10$ samples are sufficient for a reliable estimation. Furthermore, in \cref{sec:inference_time}, we demonstrate that the inference overhead is negligible. During training of GG MoE, we use a single sample to compute gradients, resulting in training times for MoE and GG MoE that are approximately the same, as illustrated in \cref{sec:time}.

\section{Datasets}
To evaluate model performance, we used $38$ publicly available datasets. Of these, $28$ were taken from \cite{grinsztajn2022tree}. These datasets are known to be more GBDT-friendly, meaning that deep learning models tend to perform worse on them. However, this collection includes only regression and binary classification problems and consists of small- to medium-sized datasets.  

To provide a more representative evaluation, we also included $10$ datasets from \cite{gorishniy2022embeddings}. This set features two multiclass classification tasks and three datasets with more than $100,000$ rows. These three datasets are also used to compare model runtime (see \cref{efficiency}).  

We provide more detailed information about each dataset in \cref{app:datasets}.

\section{Experiment Setup}

\begin{table*}[t]
\caption{Hyperparameter search space for different models.}
\label{table:hyperparameter-search}
\vskip 0.15in
\begin{center}
\begin{small}
\begin{sc}
\begin{tabular}{lcccccc}
\toprule
Model & n\_blocks & d\_block & Dropout & d\_block\_per\_expert & Tau \\
\midrule
MLP       & \( U\{1, 6, 1\} \) & \( U\{64, 1024, 16\} \)  & \( \{0,U[0, 0.5]\} \)  & N/A & N/A \\
MoE       & \( U\{1, 6, 1\} \) & \( U\{128, 1280, 64\} \)  & \( \{0,U[0, 0.5]\} \)  & \( U\{32, 64, 32\} \) & N/A \\
GG MoE & \( U\{1, 6, 1\} \) & \( U\{128, 1280, 64\} \) & \( \{0,U[0, 0.5]\} \)  & \( U\{32, 64, 32\} \) & \( U[0.5, 3] \) \\
\bottomrule
\end{tabular}
\end{sc}
\end{small}
\end{center}
\vskip -0.1in
\end{table*}

The authors of \cite{gorishniy2024tabm} provided benchmarks for a diverse set of models, including GBDT and deep learning approaches. We consider comparable benchmarks essential for evaluating models in the tabular domain. Therefore, we fully adopted their experimental setup to ensure the comparability of our results. In this section, we outline the key aspects of this setup.

\subsection{Data Preprocessing}
Binary features were mapped to $\{0,1\}$ without any additional preprocessing. For categorical features, we applied one-hot encoding. Numerical features were preprocessed using quantile normalization \cite{pedregosa2011scikit}.  

Additionally, we utilized non-linear piecewise-linear embeddings for numerical features, as proposed in \cite{gorishniy2024tabm}. The embedding dimension and the number of bins were tuned as hyperparameters (see \cref{hyperparameters}). All models were evaluated in two configurations: with and without piecewise-linear embeddings. Throughout this paper, we refer to models with embeddings by adding the prefix 'E+' to the model name (e.g., E+MoE).

\subsection{Training}
We minimized the mean squared error (MSE) loss for regression tasks and the cross-entropy loss for classification tasks. All models were trained using the AdamW optimizer \cite{loshchilov2017decoupled}. The learning rate and weight decay were tuned as hyperparameters (see \cref{hyperparameters}). We did not modify the learning rate during training. Additionally, global gradient clipping was set to $1.0$.  

Each model was trained until no improvement was observed on the validation set for 16 consecutive epochs. This early stopping criterion was applied during both hyperparameter tuning (see \cref{hyperparameters}) and the final evaluation (see \cref{evaluation}).   

\subsection{Hyperparameter Tuning}
\label{hyperparameters}

We used the Optuna package \cite{akiba2019optuna} to tune hyperparameters, setting the number of iterations to $100$ for each model. Hyperparameters were tuned using validation sets for every dataset.

In \cref{table:hyperparameter-search}, we present the search space for each hyperparameter for models without embeddings.\footnote{In the table, $U\{a,b,i\}$ represents a discrete uniform distribution from $a$ to $b$ with a step size of $i$. $U[a,b]$ denotes a continuous uniform distribution from $a$ to $b$.} For MoE-type models, we restricted expert sizes to either $32$ or $64$ hidden units. However, we allowed a wide range for the number of experts, from $2$ to $40$. The motivation behind this choice was to encourage the use of multiple weak learners rather than a few strong ones.

For GG MoE, we aimed to prevent the Gumbel-Softmax mechanism from converging to an undesirable distribution. Specifically, we sought to avoid convergence to an argmax distribution, as this would mean that only one expert contributes to the output. At the same time, we prevented convergence to a uniform distribution, as this would render the gating network meaningless and reduce the model to a Deep Ensemble \cite{lakshminarayanan2017simple}. To address this, we constrained the temperature parameter ($\tau$) to be neither too close to zero nor excessively large.

The search space was identical across datasets. When using an embedding layer, the search space remained the same, except that the maximum number of blocks was reduced to $5$. In \cref{table:optimizer-hyperparameters}, we provide the search spaces for optimizer and embedding layer hyperparameters for MoE-type models. For MLP and E+MLP models, the search space was the same except for the learning rate, which followed $U[3\mathrm{e}{-5}, 0.001]$.

\begin{table}[t]
\caption{Hyperparameter search space for optimizer and embedding parameters in MoE-based models.}
\label{table:optimizer-hyperparameters}
\vskip 0.15in
\begin{center}
\begin{small}
\begin{sc}
\begin{tabular}{lc}
\toprule
Parameter       & Search Space \\
\midrule
Learning Rate    & \( \text{log } U[3\mathrm{e}{-4}, 0.01] \) \\
Weight Decay                       & \( \{0,\text{log } U[1\mathrm{e}{-4}, 0.1]\} \) \\
Embedding Dimension & \( U\{8, 32, 4\} \) \\
Number of Bins     & \( U\{2, 128, 1\} \) \\
\bottomrule
\end{tabular}
\end{sc}
\end{small}
\end{center}
\vskip -0.1in
\end{table}

\subsection{Evaluation}
\label{evaluation}

For classification tasks, accuracy was used as the primary metric for tuning hyperparameters and evaluating final model performance on test sets. For regression tasks, the negative root mean square error (RMSE) served as the primary metric.

To rank the models, we followed the approach described in \cite{gorishniy2024tabm}, which does not count insignificant improvements as wins. Each model was trained from scratch $15$ times with tuned hyperparameters, using different random seeds. The average rank and standard deviation over these runs were then computed. Finally, we applied the ranking algorithm outlined in \cref{alg:ranking} separately to each dataset. Informally, this algorithm assigns the same rank to models where the difference between mean scores is smaller than the standard deviation.

\begin{algorithm}[tb]
   \caption{Assigning a rank to each model}
   \label{alg:ranking}
\begin{algorithmic}
   \STATE {\bfseries Input:} mean ($\mu$) and standard deviation ($\sigma$) of scores for each model
   \STATE Sort all models by mean score
   \STATE $rank_{i} \gets 1$
   \REPEAT
   \STATE Let $model_{i}$ be the first unranked model
   \STATE Assign $rank_{i}$ to $model_{i}$ and to all models with $\mu \geq \mu_{model_i} - \sigma_{model_i}$
   \STATE $rank_{i} \gets rank_{i} + 1$
   \UNTIL{All models are ranked}
\end{algorithmic}
\end{algorithm}

\section{Results}
\label{sec:results}

\begin{figure}[ht]
\vskip 0.2in
\begin{center}
\centerline{\includegraphics[width=\columnwidth]{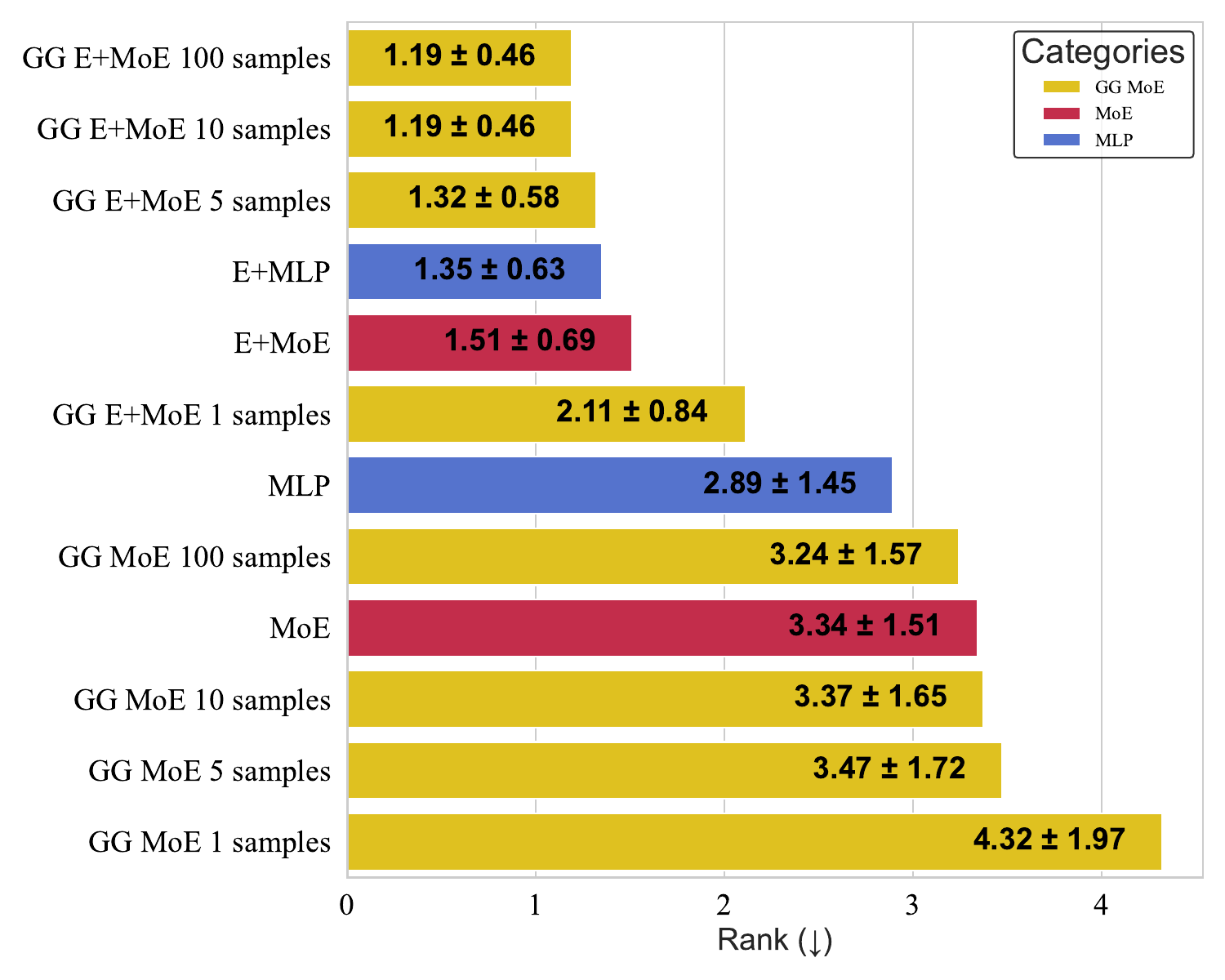}}
\caption{The average rank for each model across $38$ datasets. For each dataset, ranks were computed independently using \cref{alg:ranking}.}
\label{fig:ranks}
\end{center}
\vskip -0.2in
\end{figure}

We present the rankings for each model in \cref{fig:ranks}. For GG MoE, we applied Monte Carlo (MC) sampling (\cref{gg_moe}) using $1$, $5$, $10$, and $100$ samples. The key findings are as follows:

\begin{itemize}
    \item Models with piecewise-linear embeddings perform significantly better. This result fully aligns with the findings of \cite{gorishniy2022embeddings}. However, embeddings provide a greater benefit to MoE models, particularly GG MoE. We further discuss this in \cref{sec:reg}.
    \item Based on the scores, GG MoE is the best-performing model. However, the performance gap between GG MoE and E+MLP is small that we cannot confidently declare a significant difference between them.
    \item $10$ samples are sufficient for Monte Carlo estimation. For each dataset, using $100$ samples did not significantly improve performance compared to evaluating with $10$ samples.
\end{itemize}

\begin{table*}[t]
\caption{Statistics of the tuned model parameters across all datasets}
\label{models-hyperparameters}
\vskip 0.15in
\begin{center}
\begin{small}
\begin{sc}
\begin{tabular}{lccc ccc ccc}
\toprule
\multirow{2}{*}{Model Type} & \multicolumn{3}{c}{n\_blocks} & \multicolumn{3}{c}{d\_block} & \multicolumn{3}{c}{num\_experts} \\
& Median & Mean & Std & Median & Mean & Std & Median & Mean & Std \\
\midrule
MLP       & $4$ & $3.74$ & $1.28$ & $624$ & $564$ & $236$ & N/A & N/A & N/A \\
MoE       & $3$ & $3.37$ & $1.47$ & $704$ & $685$ & $287$ & $15$ & $16.63$ & $8.64$\\
GG MoE    & $4$ & $3.45$ & $1.51$ & $832$ & $814$ & $286$ & $18$ & $20.01$ & $11.02$  \\
E+MLP     & $3$ & $2.77$ & $1.28$ & $576$ & $609$ & $220$ & N/A & N/A & N/A \\
E+MoE     & $3$ & $2.87$ & $1.40$ & $640$ & $718$ & $327$ & $14$ & $15.54$ & $9.70$\\
GG E+MoE  & $3$ & $2.77$ & $1.15$ & $832$ & $837$ & $260$ & $16$ & $18.10$ & $9.10$ \\
\bottomrule
\end{tabular}
\end{sc}
\end{small}
\end{center}
\vskip -0.1in
\end{table*}

\section{Models Efficiency}
\label{efficiency}
\subsection{Number of Parameters}
\begin{table}[t]
\caption{Number of parameters for different model types \\ (in millions) across all datasets}
\label{model-parameters}
\vskip 0.15in
\begin{center}
\begin{small}
\begin{sc}
\begin{tabular}{lccc}
\toprule
Model Type & Median  & Mean & Std \\
\midrule
MLP        & $0.88$  & $1.19$  & $0.92$  \\
MoE        & $0.08$  & $0.12$  & $0.13$  \\
GG MoE     & $0.11$  & $0.16$  & $0.17$  \\
E+MLP      & $0.83$  & $1.22$  & $1.49$  \\
E+MoE      & $0.34$  & $1.00$  & $2.96$  \\
GG E+MoE   & $0.34$  & $0.79$  & $1.80$  \\
\bottomrule
\end{tabular}
\end{sc}
\end{small}
\end{center}
\vskip -0.1in
\end{table}

In \cref{model-parameters}, we present the average, median, and standard deviation (std) of the number of parameters (in millions) per dataset for each model.\footnote{Throughout this paper, we report both the median and average for every statistic across datasets due to the skewed nature of the distributions.} GG MoE and MoE models have approximately the same number of parameters, which is significantly lower than that of MLP models. This holds true for both models with and without embeddings.  

However, while the difference in the number of parameters between MLP and E+MLP models is negligible, the same does not apply to MoE and GG MoE models. This discrepancy arises because the embedding layer is a fully connected linear layer, which naturally leads to a significant increase in the number of parameters in MoE models.  

At the same time, the number of parameters in the backbone\footnote{The backbone refers to the subset of the model architecture excluding embedding layers.} of all models decreases when an embedding layer is added. See the number of blocks and block dimensions in \cref{models-hyperparameters}.


\subsection{Regularization}
\label{sec:reg}

In \cref{table:tau}, we observe that during hyperparameter optimization, the temperature in Gumbel-Softmax ($\tau$) was selected to encourage the contribution of every expert and introduce more stochasticity rather than enforcing sparsity. This is an interesting finding, and a possible follow-up in future work could be to increase the number of experts and examine whether $\tau$ starts to decrease.  

The only parameter that significantly differs between MoE and GG MoE models is dropout (\cref{table:dropout}). We believe this difference is primarily related to the stochasticity in the gating network, which acts as a regularization mechanism. This, in turn, leads to a lower dropout rate in the experts, resulting in stronger performance of GG MoE compared to MoE.

\begin{table}[t]
\caption{Tuned temperature parameter ($\tau$) of the Gumbel-Softmax \\ distribution}
\label{table:tau}
\vskip 0.15in
\begin{center}
\begin{small}
\begin{sc}
\begin{tabular}{lccc}
\toprule
\multirow{2}{*}{Model Type} & \multicolumn{3}{c}{tau} \\
& Median & Mean & Std \\
\midrule
GG MoE    & $2.21$ & $2.10$ & $0.62$ \\
GG E+MoE  & $2.06$ & $2.16$ & $0.57$ \\
\bottomrule
\end{tabular}
\end{sc}
\end{small}
\end{center}
\vskip -0.1in
\end{table}

\begin{table}[t]
\caption{Tuned dropout rates for different models}
\label{table:dropout}
\vskip 0.15in
\begin{center}
\begin{small}
\begin{sc}
\begin{tabular}{lccc}
\toprule
\multirow{2}{*}{Model Type} & \multicolumn{3}{c}{dropout} \\
& Median & Mean & Std \\
\midrule
MLP       & $0.165$ & $0.200$ & $0.172$ \\
MoE       & $0.167$ & $0.173$ & $0.167$ \\
GG MoE    & $0.032$ & $0.105$ & $0.148$ \\
E+MLP     & $0.209$ & $0.207$ & $0.165$ \\
E+MoE     & $0.162$ & $0.188$ & $0.177$ \\
GG E+MoE  & $0.063$ & $0.132$ & $0.153$ \\
\bottomrule
\end{tabular}
\end{sc}
\end{small}
\end{center}
\vskip -0.1in
\end{table}

\subsection{Computation Time}
\label{sec:time}
\subsubsection{Training Time}

In \cref{time-training-table}, we present the training times\footnote{Training time also includes evaluation on validation sets after each epoch.} using tuned hyperparameters for three datasets where the number of training rows exceeds $100,000$.

Both MoE-type models with embeddings outperform E+MLP. Specifically, E+MoE is significantly\footnote{Significantly means that the mean training time plus the standard deviation of E+MoE is less than the mean training time of E+MLP.} faster than E+MLP across all three datasets. GG E+MoE is significantly faster on two datasets and performs comparably to E+MLP on one dataset.
\begin{table}[t]
\caption{Mean and standard deviation of computation times (in minutes) for models on the three largest datasets}

\label{time-training-table}
\vskip 0.15in
\begin{center}
\begin{small}
\begin{sc}
\resizebox{\columnwidth}{!}{%
\begin{tabular}{lcccccc}
\toprule
Model & \multicolumn{2}{c}{covtype2} & \multicolumn{2}{c}{microsoft} & \multicolumn{2}{c}{black-friday} \\
      & Mean & Std & Mean & Std & Mean & Std \\
\midrule
E+MLP        & $9.05$ & $1.00$ & $5.82$ & $0.85$ & $1.97$ & $0.24$ \\
E+MoE        & $7.28$ & $1.23$ & $\mathbf{2.98}$ & $0.55$ & $\mathbf{1.71}$ & $0.13$ \\
GG E+MoE     & $\mathbf{5.11}$ & $0.49$ & $\mathbf{3.42}$ & $0.37$ & $2.20$ & $0.22$ \\
\midrule
MLP          & $7.33$ & $0.98$ & $\mathbf{2.58}$ & $0.39$ & $2.68$ & $0.38$ \\
MoE          & $\mathbf{5.37}$ & $1.32$ & $\mathbf{2.51}$ & $0.70$ & $\mathbf{1.29}$ & $0.20$ \\
GG MoE       & $9.95$ & $1.69$ & $3.60$ & $0.62$ & $1.53$ & $0.26$ \\
\bottomrule
\end{tabular}
}
\end{sc}
\end{small}
\end{center}
\vskip -0.1in
\end{table}

\subsubsection{GG MoE: Inference Time for Different Numbers of Samples}
\label{sec:inference_time}
As discussed in \cref{gg_moe}, the Monte Carlo (MC) estimation of the expected value does not introduce any runtime overhead during training. In \cref{table:inference}, we report the average inference time for each dataset where the number of training rows exceeds $10,000$. There are $15$ such datasets.  

For each dataset, we measured inference time using all available data, i.e., by combining the training, validation, and test sets. To reduce variance in time evaluation, we repeated the measurement $15$ times and then computed the average.  

We observed no difference in runtime between $1$, $5$, and $10$ samples, while computing $100$ samples increased inference time by approximately $33\%$. However, computing $100$ samples is unnecessary, as it does not improve accuracy (see \cref{sec:results}). This result also holds for GG MoE models without embeddings.

\begin{table}[t]
\caption{Inference time for GG E+MoE in ms.}
\label{table:inference}
\vskip 0.15in
\begin{center}
\begin{small}
\begin{sc}
\begin{tabular}{lcc}
\toprule
Model  & Mean   & Std \\
\midrule
GG E+MoE $10$ samples  & $30.69$ & $7.74$ \\
GG E+MoE $1$ samples   & $32.24$ & $8.93$ \\
GG E+MoE $5$ samples   & $32.45$ & $9.29$ \\
GG E+MoE $100$ samples & $40.61$ & $6.75$ \\
\bottomrule
\end{tabular}
\end{sc}
\end{small}
\end{center}
\vskip -0.1in
\end{table}

\section{Conclusion and Future Work}
In this paper, we compared the performance of MoE models and MLP models in the tabular domain. We introduced GG MoE, a mixture-of-experts model in which the gating network employs a Gumbel-Softmax function instead of a standard Softmax function. Our results show that this approach, combined with a piecewise-linear embedding layer, outperforms both standard MoE and MLP models.  

Additionally, we demonstrated that GG MoE and MoE models are significantly more efficient in terms of parameter count compared to MLP models, making them more suitable for scaling or ensemble-based approaches.  

We believe that this work highlights the promising potential of MoE models for tabular data in deep learning. However, there are still many avenues for further research. One direction is scaling MoE and GG MoE models, not merely by increasing the number of parameters but also by adopting more efficient ensemble techniques. Furthermore, it would be valuable to explore the performance of both well-known MoE variants, such as Hierarchical MoE, and emerging deep learning architectures, such as sparse or soft MoE.

\section*{Impact Statement}
This paper presents work whose goal is to advance the field of 
Machine Learning. There are many potential societal consequences 
of our work, none which we feel must be specifically highlighted here.


\bibliography{main}
\bibliographystyle{icml2025}

\newpage
\appendix
\onecolumn
\section{Datasets Overview}
\label{app:datasets}
In \cref{table:datasets}, we present the statistics for each dataset used in the evaluation. The first $28$ datasets were taken from \cite{gorishniy2022embeddings}, the last $10$ were sourced from \cite{grinsztajn2022tree}.

\begin{table*}[t]
\caption{Main Datasets characteristics.}
\label{table:datasets}
\vskip 0.15in
\begin{center}
\begin{small}
\begin{sc}
\resizebox{\textwidth}{!}{%
\begin{tabular}{lcccccccc}
\toprule
ID & Task Type & n\_num\_features & n\_cat\_features & Train Size & Val Size & Test Size & n\_bin\_features & n\_classes \\
\midrule
wine & binclass & 11 & - & 1787 & 230 & 537 & - & - \\
phoneme & binclass & 5 & - & 2220 & 285 & 667 & - & - \\
analcatdata\_supreme & regression & 2 & - & 2836 & 364 & 852 & 5 & - \\
Mercedes\_Benz\_Greener\_Manufacturing & regression & 0 & 3 & 2946 & 378 & 885 & 356 & - \\
KDDCup09\_upselling & binclass & 34 & 14 & 3589 & 461 & 1078 & 1 & - \\
kdd\_ipums\_la\_97-small & binclass & 20 & - & 3631 & 467 & 1090 & - & - \\
wine\_quality & regression & 11 & - & 4547 & 585 & 1365 & - & - \\
isolet & regression & 613 & - & 5457 & 702 & 1638 & - & - \\
cpu\_act & regression & 21 & - & 5734 & 737 & 1721 & - & - \\	
bank-marketing & binclass & 7 & - & 7404 & 952 & 2222 & - & - \\
Brazilian\_houses & regression & 8 & 1 & 7484 & 962 & 2246 & 2 & - \\
MagicTelescope & binclass & 10 & - & 9363 & 1203 & 2810 & - & - \\
Ailerons & regression & 33 & - & 9625 & 1237 & 2888 & - & - \\
MiamiHousing2016 & regression & 13 & - & 9752 & 1254 & 2926 & - & - \\
OnlineNewsPopularity & regression & 45 & - & 9999 & 8893 & 20752 & 14 & - \\
elevators & regression & 16 & - & 10000 & 1979 & 4620 & - & - \\
credit & binclass & 10 & - & 10000 & 2014 & 4700 & - & - \\
pol & regression & 26 & - & 10000 & 1500 & 3500 & - & - \\
superconduct & regression & 79 & - & 10000 & 3378 & 7885 & - & - \\
house\_sales & regression & 15 & - & 10000 & 3483 & 8130 & 2 & - \\
medical\_charges & regression & 3 & - & 10000 & 45919 & 50000 & - & - \\
fifa & regression & 5 & - & 10000 & 2418 & 5645 & - & - \\
jannis & binclass & 54 & - & 40306 & 5182 & 12092 & - & - \\
road-safety & binclass & 29 & 3 & 50000 & 18528 & 43234 & - & - \\
particulate-matter-ukair-2017 & regression & 3 & 3 & 50000 & 50000 & 50000 & - & - \\
taxi-green-dec-2016 & regression & 9 & 4 & 50000 & 50000 & 50000 & 3 & - \\
MiniBooNE & binclass & 50 & - & 50000 & 6899 & 16099 & - & - \\
year & regression & 90 & - & 50000 & 15489 & 36141 & - & - \\
Churn\_Modelling & binclass & 7 & 1 & 6400 & 1600 & 2000 & 3 & - \\
California\_Housing & regression & 8 & 0 & 13209 & 3303 & 4128 & 0 & - \\
House\_16H & regression & 16 & 0 & 14581 & 3646 & 4557 & 0 & - \\
Adult & binclass & 6 & 7 & 26048 & 6513 & 16281 & 1 & - \\
Diamond\_OPENML & regression & 6 & 3 & 34521 & 8631 & 10788 & 0 & - \\
Otto\_Group\_Products & multiclass & 93 & 0 & 39601 & 9901 & 12376 & 0 & 9 \\
higgs-small & binclass & 28 & 0 & 62751 & 15688 & 19610 & - & - \\
black-friday & regression & 4 & 4 & 106764 & 26692 & 33365 & 1 & - \\
covtype2 & multiclass & 10 & 1 & 371847 & 92962 & 116203 & 4 & 7 \\
microsoft & regression & 131 & 0 & 723412 & 235259 & 241521 & 5 & - \\
\bottomrule
\end{tabular}
}
\end{sc}
\end{small}
\end{center}
\vskip -0.1in
\end{table*}


\end{document}